\documentclass[10pt,twocolumn,letterpaper]{article}

\usepackage[pagenumbers]{cvpr}      \usepackage{multirow}

\usepackage{placeins}
\usepackage{tabularx}
\usepackage{colortbl}
\usepackage{bm}
\definecolor{cvprblue}{rgb}{0.21,0.49,0.74}
\colorlet{tablebest}{red!40}
\colorlet{tablesecond}{orange!40}
\colorlet{tablethird}{yellow!40}
\newcommand{\tabfirst}[1]{\cellcolor{tablebest}#1}
\newcommand{\tabsecond}[1]{\cellcolor{tablesecond}#1}
\newcommand{\tabthird}[1]{\cellcolor{tablethird}#1}
\usepackage[breaklinks,colorlinks,allcolors=cvprblue]{hyperref}

\title{Learning Continuous Scene Depth Representations via Cross-Modal Gaussian Rendering}

\author{Zhiwei Zhong \quad Lingyu Zhu \quad Baoliang Chen \quad Peilin Chen \quad Shiqi Wang\\
City University of Hong Kong}
\date{}

\begin{document}
\maketitle
\begin{abstract}
Depth super-resolution with color guidance is a fundamental problem in geometric scene understanding that aims to reconstruct high-resolution depth using structural cues from color images. In practical RGB-D systems, the resolution gap between depth and color images often varies, requiring flexible upsampling factors. However, many existing methods are designed for predefined integer scales. Although implicit methods support arbitrary-scale reconstruction, they typically rely on computationally demanding per-pixel neural decoding and may struggle to recover sharp depth boundaries and fine structures. To this end, we propose a cross-modal Gaussian rendering framework that learns continuous depth representations for arbitrary-scale depth super-resolution. Specifically, we first construct hierarchical Gaussian representations for progressive depth reconstruction, with explicit spatial support defined by learnable Gaussian primitives. Building on this representation, we design a Structure-Aware Gaussian Aggregation module that combines spatial weights with cross-modal feature affinities to reduce aggregation across depth discontinuities. We then develop a Scale-Adaptive Gaussian Rendering module that adapts Gaussian spatial kernels to the target pixel footprint. To further facilitate cross-modal learning, we propose a training strategy that gradually introduces higher-frequency guidance information during training. Extensive experiments on benchmark datasets demonstrate the effectiveness of the proposed method at both in-distribution and out-of-distribution upsampling scales.

\end{abstract}

\section{Introduction}
\begin{figure}[!t]
    \centering
    \includegraphics[width=\linewidth]{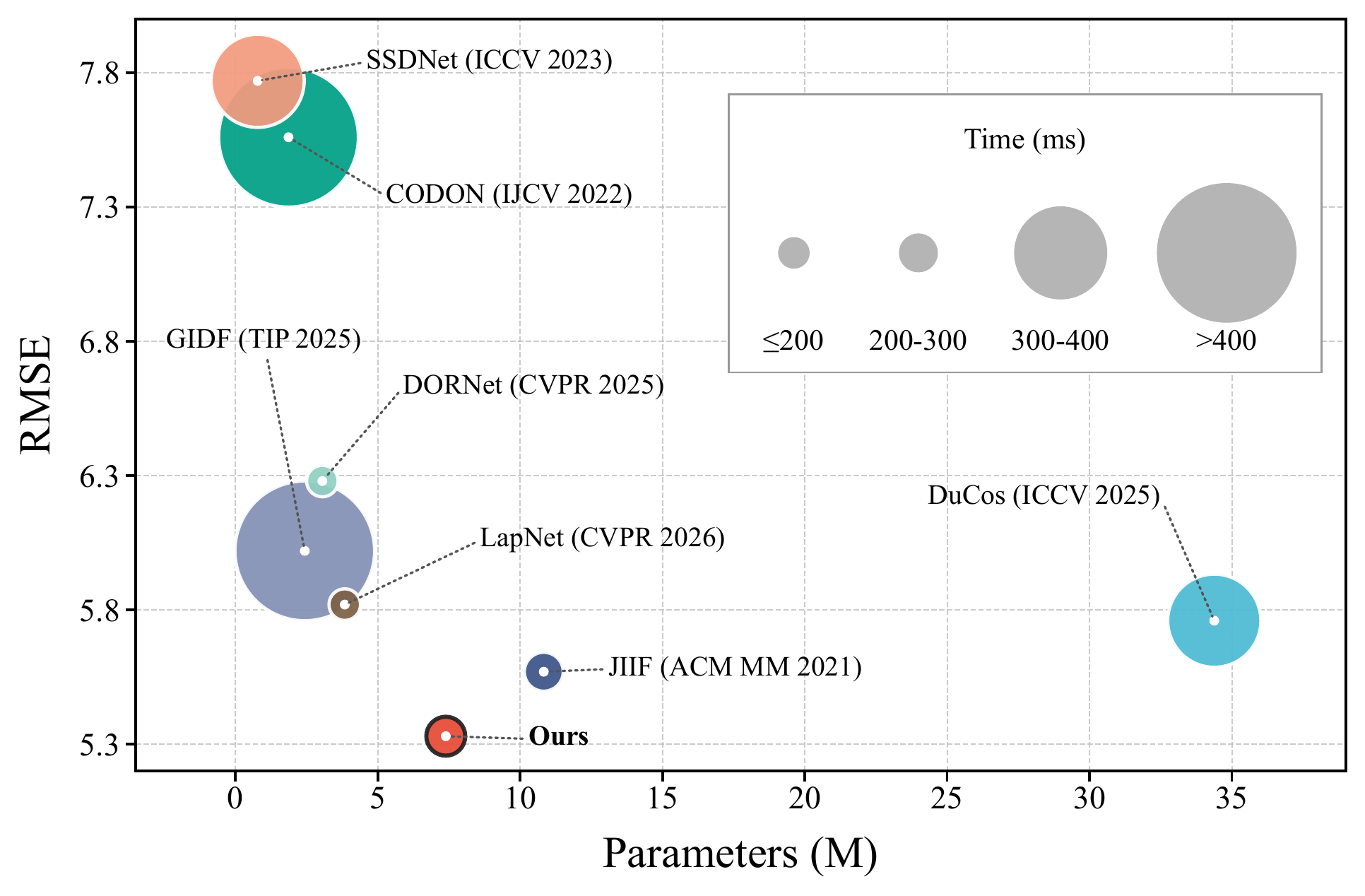}
    \caption{Quantitative comparison on the RGB-D-D dataset for $\times16$ guided depth map super-resolution. Inference latency is measured on an NVIDIA 3090 GPU using $60\times60$ LR depth inputs.}
    \label{fig:efficiency}
\end{figure}

Depth maps encode the distance from the camera to scene surfaces at each pixel and provide geometric cues for applications such as 3D reconstruction, visual SLAM, and scene understanding~\cite{gupta2014learning,newcombe2011kinectfusion,tateno2017cnnslam}. However, hardware limitations and acquisition constraints often result in low-resolution depth maps, which can compromise the performance of downstream applications~\cite{yan2025tpvd}. A common approach to addressing this problem is guided depth map super-resolution (GDSR), a technique that reconstructs a high-resolution (HR) depth map from its low-resolution (LR) observation with the assistance of a paired HR RGB image~\cite{zhong2023gdsr_survey,qiao2026efficient}.

Learning-based GDSR methods have improved depth reconstruction by modeling cross-modal interactions between RGB images and depth maps~\cite{metzger2023dad,zhao2022dctnet,zhao2023ssdnet,kang2025c2pd,wang2025dornet,yan2025ducos,dong2026robself}. Nevertheless, many existing methods are designed for predefined integer upsampling factors and often require separate models to support different factors. Such scale-specific designs limit flexibility in accommodating varying resolution ratios across RGB-D devices and the output resolutions required by downstream tasks. Implicit neural representations offer a flexible solution by modeling depth as a continuous function of spatial coordinates. For example, JIIF formulates GDSR as neural implicit interpolation and jointly predicts interpolation weights and values~\cite{tang2021jiif}. GeoDSR introduces a scale-modulated distance field into a geometric spatial aggregator to model depth continuously across spatial locations and scales~\cite{wang2023geodsr}. GIDF further incorporates scale information into feature encoding and decoding through cross-modal fusion and implicit attention~\cite{zuo2025gidf}.

Despite this flexibility, implicit methods still face challenges in computational efficiency and geometric detail recovery. First, existing methods rely on per-pixel neural decoding to reconstruct depth from neighboring features, leading to increasing computational costs as the output resolution grows. Second, local feature aggregation may mix information from different surfaces near depth discontinuities, making fine geometric details difficult to preserve. These challenges motivate exploring an explicit representation that supports continuous depth reconstruction with controllable spatial support and structure-aware aggregation.

Recent advances in arbitrary-scale image SR suggest that 2D Gaussian primitives provide a suitable representation for this purpose~\cite{chen2025gsasr,hu2024gaussiansr,peng2025pixel2gaussian}. These methods represent an image or its features as a collection of learnable Gaussian primitives whose spatial support adapts to local image structure. Output pixel values are reconstructed by combining Gaussian contributions at the query locations, so the same representation can be evaluated on different grids to produce images at different resolutions. This explicit formulation provides a promising basis for arbitrary-scale depth reconstruction. However, directly applying Gaussian rendering to GDSR still presents several challenges.

First, spatial proximity alone cannot determine whether Gaussian primitives lie on the same depth surface. Aggregating nearby primitives across a depth discontinuity can blur object boundaries. Guidance provides useful structural cues, but its appearance variations do not always correspond to depth changes. Reliable aggregation therefore requires jointly considering depth and RGB structure. Second, each output pixel covers a spatial region whose size depends on the requested resolution. Over the same image domain, a coarse-grid pixel covers a larger area than a fine-grid pixel. Evaluating the Gaussian response only at the pixel center ignores this difference. The spatial support used for rendering should therefore adapt to the target pixel footprint.

To address these challenges, we propose a cross-modal Gaussian rendering framework for arbitrary-scale GDSR. The framework predicts hierarchical Gaussian representations from the input and progressively reconstructs depth by combining their rendered maps. To preserve geometric boundaries, we develop Structure-Aware Gaussian Aggregation (SAGA), which combines spatial weights with cross-modal feature affinities to reduce the influence of primitives across depth discontinuities. To accommodate the spatial coverage of output pixels, we introduce Scale-Adaptive Gaussian Rendering (SAGR), which incorporates the target pixel footprint into Gaussian kernels and adjusts their amplitudes to preserve each kernel’s integral. Finally, to facilitate cross-modal learning, we introduce a progressive RGB guidance training strategy that gradually incorporates finer RGB details by reducing guidance blur during training. The main contributions are summarized as follows:
\begin{itemize}
    \item We propose a cross-modal Gaussian rendering framework for arbitrary-scale GDSR that progressively reconstructs depth using hierarchical Gaussian representations.
    \item We develop a Structure-Aware Gaussian Aggregation (SAGA) mechanism that combines spatial weights with cross-modal affinities to reduce aggregation across depth discontinuities and preserve geometric boundaries.
    \item We propose a Scale-Adaptive Gaussian Rendering (SAGR) scheme that incorporates the target pixel footprint into Gaussian spatial kernels, adapting their support to the sampling area of each output pixel.
    \item We introduce a progressive training strategy that facilitates cross-modal learning by gradually reducing guidance blur. Experiments on both datasets demonstrate accurate arbitrary-scale reconstruction and a favorable trade-off between accuracy and model size (Fig.~\ref{fig:efficiency}).
\end{itemize}

\section{Related Work}

\textbf{Guided Depth Super-Resolution.}
Guided depth super-resolution exploits the structural correspondence between RGB images and depth maps to recover spatial details missing from low-resolution depth observations. Existing approaches can be broadly categorized into filtering-based, optimization-based, and learning-based methods~\cite{zhong2023gdsr_survey}.

Filtering-based methods use spatial proximity and guidance information to determine local filtering weights~\cite{kopf2007joint_bilateral_upsampling,he2013guidedfilter}. These methods are efficient and require no training, but discrepancies between RGB appearance and depth geometry can lead to texture transfer or boundary blurring. Optimization-based methods combine observation fidelity with image-guided geometric regularization~\cite{diebel2006mrf,ferstl2013tgv}. They explicitly enforce data consistency and spatial regularity, but depend on the chosen priors and often require iterative optimization. Learning-based methods learn cross-modal fusion or adaptive filtering from data~\cite{hui2016msgnet,kim2019dkn,zhao2022dctnet,zhao2023ssdnet,pan2019svlrm,deng2021cunet,yan2022mmnet}. Recent approaches further incorporate degradation-aware fusion and foundation-model guidance to better exploit cross-modal information~\cite{wang2025dornet,yan2025ducos}. Despite these advances, most existing methods are designed for predefined upsampling factors, limiting their flexibility when input resolution ratios or target output sizes vary across RGB-D systems and downstream visual perception tasks.

\textbf{Arbitrary-Scale Super-Resolution.}
Arbitrary-scale SR reconstructs images at different resolutions using a unified model. Meta-SR predicts scale-dependent upsampling filters~\cite{hu2019metasr}, LIIF learns continuous image representations through local implicit functions~\cite{wang2021liif}, and LTE incorporates local frequency information to improve texture recovery~\cite{lee2022lte}. CLIT combines cross-scale local attention and frequency encoding with a cascaded architecture to exploit multi-scale features for progressive reconstruction~\cite{chen2023clit}.

For GDSR, JIIF introduces guided implicit interpolation~\cite{tang2021jiif}, GeoDSR incorporates scale-modulated geometric aggregation~\cite{wang2023geodsr}, and GIDF develops scale-aware cross-modal feature fusion and implicit decoding~\cite{zuo2025gidf}. Although these methods enable continuous depth reconstruction, they may struggle to recover fine geometric details. Moreover, repeated neural decoding at query locations can increase computational costs at high output resolutions.

\textbf{Gaussian-Based Image Reconstruction.}
Anisotropic Gaussian primitives represent 3D scenes and enable real-time novel-view synthesis through differentiable splatting~\cite{kerbl2023gaussians}. To improve surface reconstruction, Huang \textit{et~al.}~\cite{huang2024twodgs} introduce 2D Gaussian splatting with oriented planar Gaussian disks embedded in 3D space. For image restoration, Gaussian primitives are instead defined directly in the 2D image domain to represent images or their features. GaussianSR~\cite{hu2024gaussiansr} employs continuous Gaussian feature representations for arbitrary-scale SR, while GSASR~\cite{chen2025gsasr} and Pixel to Gaussian~\cite{peng2025pixel2gaussian} develop efficient Gaussian prediction and rendering methods for continuous image reconstruction. These image-domain methods aggregate Gaussian responses on query grids, allowing a single set of predicted primitives to be reused for image reconstruction at different output resolutions during inference.

\begin{figure*}[!t]
    \centering
    \includegraphics[width=\textwidth]{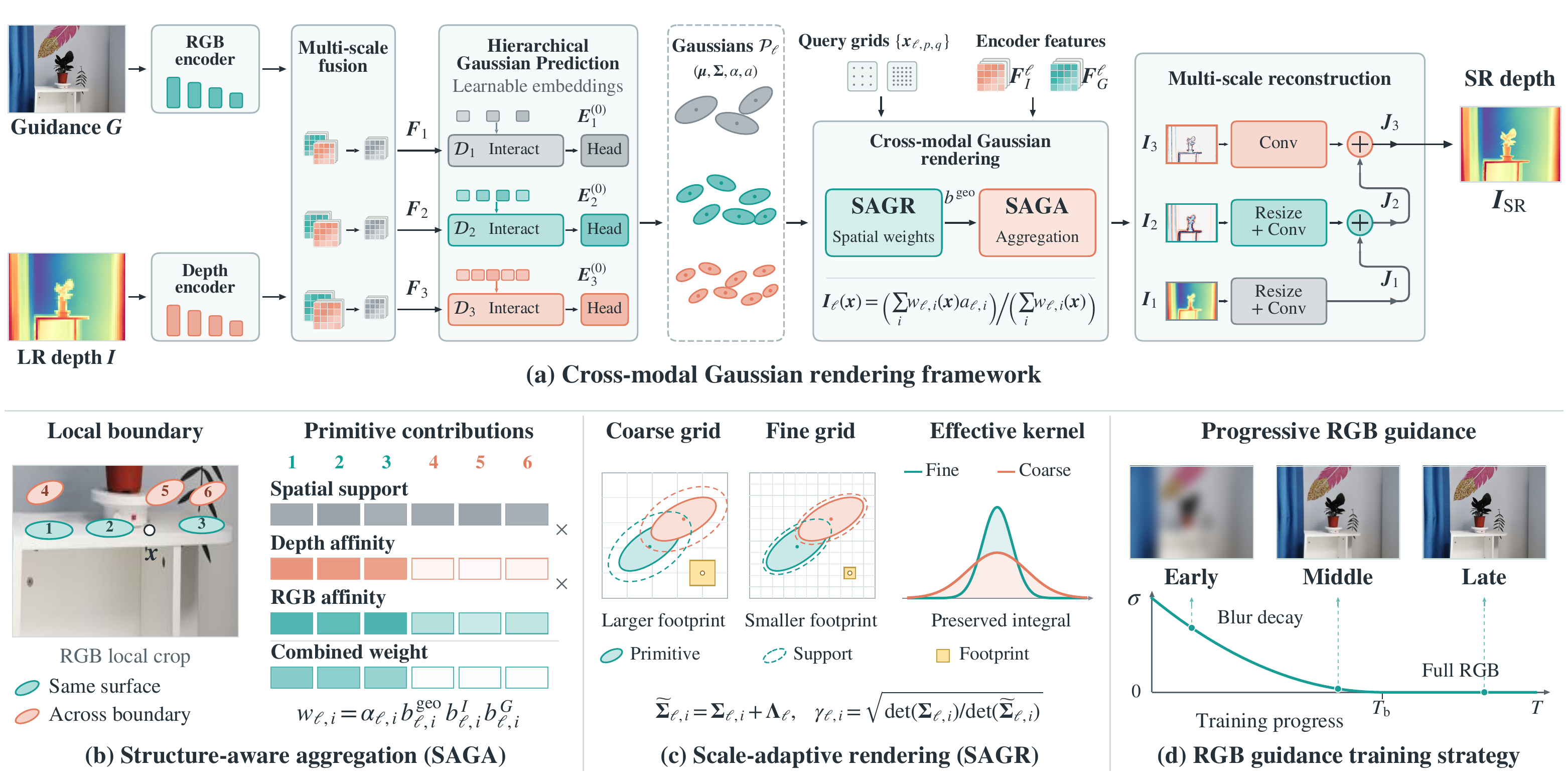}
    \caption{Overview of the proposed cross-modal Gaussian rendering framework.
    (a) Fused RGB and depth features predict hierarchical Gaussian primitives for rendering and multi-scale reconstruction.
    (b) Structure-Aware Gaussian Aggregation (SAGA) combines spatial weights with cross-modal feature affinities to preserve depth boundaries.
    (c) Scale-Adaptive Gaussian Rendering (SAGR) adapts Gaussian kernels to the target pixel footprint.
    (d) Progressive RGB guidance introduces finer color details by reducing guidance blur during training.}
    \label{fig:framework}
\end{figure*}

\section{Method}

\subsection{Problem Formulation and Overview}
\label{sec:overview}

Given a low-resolution (LR) depth map $\boldsymbol{I}_{\mathrm{LR}}\in\mathbb{R}^{h\times w}$ and a high-resolution (HR) guidance image $\boldsymbol{G}\in\mathbb{R}^{H\times W\times3}$, guided depth super-resolution (GDSR) aims to recover an HR depth map $\boldsymbol{I}_{\mathrm{SR}}\in\mathbb{R}^{H_{\mathrm{t}}\times W_{\mathrm{t}}}$ using structural cues from $\boldsymbol{G}$. In this work, we focus on arbitrary-scale reconstruction where the upsampling factor is not restricted to predefined values. To facilitate subsequent feature fusion, we first upsample $\boldsymbol I_{\mathrm{LR}}$ to the resolution of $\boldsymbol G$, obtaining $\boldsymbol I\in\mathbb R^{H\times W}$.

Building on this formulation, we develop a cross-modal Gaussian rendering framework, as illustrated in Fig.~\ref{fig:framework}. Two separate encoders extract multi-scale features from $\boldsymbol I$ and $\boldsymbol G$, which are fused at corresponding levels to obtain multi-scale features $\{\boldsymbol F_\ell\}_{\ell=1}^{N}$, where $N$ denotes the number of levels. The fused features interact with learnable Gaussian embeddings to predict hierarchical Gaussian primitives. Scale-Adaptive Gaussian Rendering (SAGR) and Structure-Aware Gaussian Aggregation (SAGA) then render a depth map $\boldsymbol I_\ell$ at each level. Finally, we combine the rendered maps through multi-scale reconstruction to obtain HR depth map:
\begin{equation}
\boldsymbol{I}_{\mathrm{SR}}
=\sum_{\ell=1}^{N}\mathcal{C}_\ell\!\left(\mathcal{R}_\ell(\boldsymbol{I}_\ell)\right),
\label{eq:multiscale_reconstruction}
\end{equation}
where $\mathcal{R}_\ell$ and $\mathcal{C}_\ell$ denote resizing and convolution operations, respectively. The reconstructed depth map is supervised by $L_1$ loss against the HR depth map.

\subsection{Hierarchical Gaussian Representation}
\label{sec:continuous}
\label{sec:representation}

We represent scene depth using hierarchical sets of anisotropic 2D Gaussian primitives. We first define the continuous depth representation and then describe how Gaussian parameters are predicted from the fused features.

\textbf{Gaussian representation.} We describe a single level and omit its index for simplicity. A 2D Gaussian describes a local spatial response on the image plane. Its center $\boldsymbol{\mu}_i\in\mathbb{R}^{2}$ specifies its location, while its positive-definite covariance $\boldsymbol{\Sigma}_i\in\mathbb{R}^{2\times2}$ determines the shape of its spatial support. At a normalized image coordinate $\boldsymbol{x}\in[0,1]^2$, the Gaussian spatial response is computed as follows:
\begin{equation}
g_i(\boldsymbol{x})=\exp\!\left[-\frac12(\boldsymbol{x}-\boldsymbol{\mu}_i)^{\top}\boldsymbol{\Sigma}_i^{-1}(\boldsymbol{x}-\boldsymbol{\mu}_i)\right].
\label{eq:gaussian_response}
\end{equation}
Each Gaussian is defined on the image plane and carries a scalar depth attribute $a_i$ and a contribution strength $\alpha_i$. A collection of $M$ primitives forms the representation $\mathcal{P}=\{(\boldsymbol{\mu}_i,\boldsymbol{\Sigma}_i,\alpha_i,a_i)\}_{i=1}^{M}$. A basic continuous depth field is defined by normalized Gaussian aggregation:
\begin{equation}
I(\boldsymbol{x})=
\frac{\sum_{i=1}^{M}\alpha_i g_i(\boldsymbol{x})a_i}
{\sum_{i=1}^{M}\alpha_i g_i(\boldsymbol{x})}.
\label{eq:continuous_depth}
\end{equation}
This continuous field can be queried at arbitrary spatial locations for depth reconstruction at different resolutions.

\textbf{Gaussian prediction.} To predict these representations from the input, we adopt a window-based decoder at each level. The fused features $\boldsymbol F_\ell$ are projected to the embedding dimension and partitioned into non-overlapping local windows. Each window is assigned a copy of a learnable embedding table $\boldsymbol E_\ell^{(0)}$. The initial embeddings are shared across windows within the same level and updated using the fused features in their corresponding local windows.

The Gaussian embeddings act as queries in cross-attention, while the local fused features provide keys and values. Self-attention with alternating regular and shifted windows then exchanges information among the updated embeddings. These operations and residual connections form the decoder $\mathcal D_\ell$ and produce the Gaussian features:
\begin{equation}
\boldsymbol Z_\ell=\mathcal D_\ell\!\left(\boldsymbol E_\ell^{(0)},\boldsymbol F_\ell\right).
\label{eq:hierarchical_decoding}
\end{equation}
Each updated embedding predicts one Gaussian primitive through separate MLP heads for the center offset, standard deviations, correlation coefficient, contribution strength, and scalar depth attribute. The predicted offset is scaled by the reference-grid spacing and added to the corresponding reference position. The standard deviations and correlation coefficient determine the covariance. The Gaussian centers and resulting primitive set are given by:
\begin{equation}
\boldsymbol\mu_{\ell,i}=\boldsymbol p_{\ell,i}+\Delta\boldsymbol\mu_{\ell,i},\qquad
\mathcal P_\ell=\mathcal H_\ell(\boldsymbol Z_\ell),
\label{eq:parameter_prediction}
\end{equation}
where the reference positions $\boldsymbol p_{\ell,i}$ form a uniform grid, and the offsets $\Delta\boldsymbol\mu_{\ell,i}$ are expressed in normalized image coordinates. The parameter prediction mapping $\mathcal H_\ell$ comprises the MLP heads and the conversion to Gaussian parameters. Each level uses independently parameterized decoders and prediction heads to construct its Gaussian representation.

\subsection{Cross-Modal Gaussian Rendering}
\label{sec:rendering}

We render each Gaussian set through scale-adaptive spatial weighting and structure-aware aggregation. At each level, SAGR computes spatial weights, and SAGA combines them with depth and RGB feature affinities. These weights aggregate depth attributes into a rendered map.

\subsubsection{Scale-Adaptive Gaussian Rendering}
\label{sec:sagr}

SAGR adapts Gaussian spatial support to the pixel footprint as the rendering scale changes, as illustrated in Fig.~\ref{fig:framework}(c). For a rendering grid of size $H_\ell\times W_\ell$, each pixel covers a rectangle of width $1/W_\ell$ and height $1/H_\ell$ in the normalized image domain. Higher rendering resolutions produce smaller pixel footprints. We approximate this footprint by a normalized Gaussian with matching covariance:
\begin{equation}
\boldsymbol{\Lambda}_\ell=\frac1{12}
\begin{bmatrix}W_\ell^{-2}&0\\0&H_\ell^{-2}\end{bmatrix}.
\label{eq:pixel_footprint}
\end{equation}
Convolving each unit-peak Gaussian kernel with the normalized footprint accounts for the spatial coverage of the rendered pixel. The Gaussian convolution yields the following effective covariance and amplitude correction:
\begin{equation}
\begin{aligned}
\widetilde{\boldsymbol{\Sigma}}_{\ell,i}
&=\boldsymbol{\Sigma}_{\ell,i}+\boldsymbol{\Lambda}_\ell,\\
\gamma_{\ell,i}&=\sqrt{\frac{\det(\boldsymbol{\Sigma}_{\ell,i})}
{\det(\widetilde{\boldsymbol{\Sigma}}_{\ell,i})}}.
\end{aligned}
\label{eq:effective_covariance}
\end{equation}
The resulting Gaussian kernel defines the spatial weight for depth aggregation at each query location:
\begin{equation}
b_{\ell,i}^{\mathrm{geo}}(\boldsymbol{x})
=\gamma_{\ell,i}\exp\!\left[-\frac12
(\boldsymbol{x}-\boldsymbol{\mu}_{\ell,i})^{\top}
\widetilde{\boldsymbol{\Sigma}}_{\ell,i}^{-1}
(\boldsymbol{x}-\boldsymbol{\mu}_{\ell,i})\right].
\label{eq:spatial_contribution}
\end{equation}
A coarser grid produces broader effective kernels, while a finer grid reduces footprint-induced smoothing. The amplitude factor $\gamma_{\ell,i}$ preserves the integral of each spatial kernel under the Gaussian footprint approximation.
\begin{table*}[!t]
\centering\footnotesize
\setlength{\tabcolsep}{1.5pt}\renewcommand{\arraystretch}{1.08}
\caption{Quantitative comparison on the TOFDSR dataset~\cite{yan2025tpvd} for arbitrary-scale RGB-guided depth super-resolution. The best, second-best, and third-best results in each column are highlighted with \colorbox{tablebest}{\strut red}, \colorbox{tablesecond}{\strut orange}, and \colorbox{tablethird}{\strut yellow} backgrounds, respectively.}
\label{tab:tof_metrics}
\begin{tabularx}{\textwidth}{@{}l*{18}{>{\centering\arraybackslash}X}@{}}
\toprule
\multirow{3}{*}{Methods} & \multicolumn{9}{c}{In-distribution} & \multicolumn{9}{c}{Out-of-distribution} \\
\cmidrule(lr){2-10}\cmidrule(lr){11-19}
 & \multicolumn{3}{c}{$\times1$} & \multicolumn{3}{c}{$\times7.5$} & \multicolumn{3}{c}{$\times16$} & \multicolumn{3}{c}{$\times17.5$} & \multicolumn{3}{c}{$\times20$} & \multicolumn{3}{c}{$\times24$} \\
\cmidrule(lr){2-4}\cmidrule(lr){5-7}\cmidrule(lr){8-10}\cmidrule(lr){11-13}\cmidrule(lr){14-16}\cmidrule(lr){17-19}
 & \scriptsize RMSE & \scriptsize MAE & $\delta_{1.05}$ & \scriptsize RMSE & \scriptsize MAE & $\delta_{1.05}$ & \scriptsize RMSE & \scriptsize MAE & $\delta_{1.05}$ & \scriptsize RMSE & \scriptsize MAE & $\delta_{1.05}$ & \scriptsize RMSE & \scriptsize MAE & $\delta_{1.05}$ & \scriptsize RMSE & \scriptsize MAE & $\delta_{1.05}$ \\
\midrule
DJFR~\cite{li2019djfr} & 6.27 & 2.74 & 94.62 & 6.84 & 3.27 & 91.80 & 8.87 & 4.85 & 84.38 & 9.12 & 5.04 & 83.48 & 9.68 & 5.46 & 81.66 & 11.16 & 6.58 & 77.24 \\
SVLRM~\cite{pan2019svlrm} & 6.01 & 2.51 & 95.06 & 6.37 & 2.86 & 93.33 & 8.60 & 4.60 & 85.01 & 8.87 & 4.81 & 84.02 & 9.51 & 5.29 & 81.91 & 11.20 & 6.59 & 76.78 \\
DKN~\cite{kim2019dkn} & 5.53 & 2.46 & 96.41 & 5.74 & 2.72 & 95.14 & 7.23 & 3.95 & 89.32 & 7.49 & 4.16 & 88.28 & 8.12 & 4.65 & 85.95 & 10.03 & 6.09 & 79.74 \\
CUNet~\cite{deng2021cunet} & 5.74 & 2.54 & 96.36 & 6.03 & 2.85 & 94.87 & 7.89 & 4.29 & 88.31 & 8.18 & 4.51 & 87.21 & 8.89 & 5.04 & 84.83 & 10.85 & 6.54 & 78.63 \\
JIIF~\cite{tang2021jiif} & \tabsecond{4.81} & \tabsecond{1.76} & \tabsecond{97.05} & \tabsecond{4.80} & \tabsecond{1.86} & \tabsecond{96.70} & \tabsecond{5.48} & \tabsecond{2.45} & \tabsecond{94.16} & \tabsecond{5.69} & \tabsecond{2.64} & \tabsecond{93.25} & \tabsecond{6.27} & \tabsecond{3.13} & \tabsecond{90.78} & 9.25 & \tabthird{5.27} & 80.94 \\
FDSR~\cite{he2021rgbdd} & 5.60 & 2.30 & 95.93 & 5.77 & 2.51 & 94.89 & 7.22 & 3.69 & 89.38 & 7.48 & 3.89 & 88.41 & 8.09 & 4.36 & 86.29 & 9.93 & 5.77 & 80.21 \\
CODON~\cite{yang2022codon} & 5.44 & 2.04 & 96.20 & 5.53 & 2.20 & 95.41 & 6.56 & 2.99 & 91.86 & 6.85 & 3.21 & 90.76 & 7.58 & 3.76 & 88.10 & 10.23 & 5.71 & 80.04 \\
DCTNet~\cite{zhao2022dctnet} & 5.85 & 2.37 & 95.40 & 6.10 & 2.65 & 94.13 & 8.10 & 4.23 & 86.80 & 8.37 & 4.44 & 85.81 & 8.98 & 4.91 & 83.69 & 10.68 & 6.22 & 78.31 \\
MMNet~\cite{yan2022mmnet} & 5.66 & 2.28 & 95.99 & 5.83 & 2.50 & 94.91 & 7.41 & 3.77 & 88.98 & 7.68 & 3.98 & 87.93 & 8.29 & 4.45 & 85.84 & 10.13 & 5.85 & 79.92 \\
SSDNet~\cite{zhao2023ssdnet} & 5.40 & 2.10 & 96.32 & 5.82 & 2.49 & 94.20 & 7.42 & 3.77 & 88.51 & 7.69 & 3.98 & 87.50 & 8.37 & 4.48 & 85.29 & 10.39 & 6.00 & 79.25 \\
GeoDSR~\cite{wang2023geodsr} & 5.17 & 1.93 & 96.67 & 5.09 & 2.04 & 96.07 & 5.81 & 2.65 & 93.28 & 6.02 & 2.83 & 92.33 & 6.58 & 3.28 & 90.01 & \tabsecond{9.13} & \tabsecond{5.16} & \tabsecond{81.69} \\
GIDF~\cite{zuo2025gidf} & 5.31 & 1.96 & 96.39 & 5.29 & 2.04 & 95.92 & 6.03 & 2.65 & 93.34 & 6.22 & 2.81 & 92.59 & 6.77 & 3.27 & 90.24 & \tabthird{9.24} & \tabsecond{5.16} & \tabthird{81.67} \\
SFINet~\cite{zhou2025sfinet} & 5.40 & 2.11 & 96.17 & 5.58 & 2.30 & 95.23 & 6.87 & 3.38 & 90.30 & 7.11 & 3.59 & 89.30 & 7.69 & 4.04 & 87.23 & 9.54 & 5.46 & 81.09 \\
DORNet~\cite{wang2025dornet} & 5.25 & 1.96 & 96.55 & 5.25 & 2.08 & 95.99 & 6.08 & 2.77 & 92.73 & 6.31 & 2.97 & 91.68 & 6.99 & 3.51 & 88.94 & 9.74 & 5.53 & 80.28 \\
DuCos~\cite{yan2025ducos} & \tabthird{4.97} & \tabthird{1.81} & \tabthird{96.73} & \tabthird{4.93} & \tabthird{1.89} & \tabthird{96.35} & \tabthird{5.68} & \tabthird{2.49} & \tabthird{93.84} & \tabthird{5.92} & \tabthird{2.67} & \tabthird{92.98} & \tabthird{6.55} & \tabthird{3.16} & \tabthird{90.58} & 9.71 & 5.39 & 81.02 \\
LapNet~\cite{zhong2026lapnet} & 5.03 & 1.82 & 96.66 & 5.00 & 1.91 & 96.28 & 5.77 & 2.53 & 93.76 & 6.01 & 2.71 & 92.86 & 6.63 & 3.19 & 90.42 & 9.71 & 5.39 & 80.96 \\
\addlinespace[2pt]
\textbf{Ours} & \tabfirst{4.58} & \tabfirst{1.60} & \tabfirst{97.17} & \tabfirst{4.52} & \tabfirst{1.65} & \tabfirst{97.00} & \tabfirst{5.13} & \tabfirst{2.16} & \tabfirst{95.08} & \tabfirst{5.31} & \tabfirst{2.32} & \tabfirst{94.33} & \tabfirst{5.83} & \tabfirst{2.75} & \tabfirst{92.19} & \tabfirst{8.82} & \tabfirst{4.91} & \tabfirst{82.44} \\
\bottomrule
\end{tabularx}
\end{table*}

\subsubsection{Structure-Aware Gaussian Aggregation}
\label{sec:saga}

SAGA evaluates structural compatibility using both depth and RGB features, as shown in Fig.~\ref{fig:framework}(b). Spatially nearby primitives may lie on different surfaces, and RGB appearance changes do not always indicate depth discontinuities. We therefore compare query and primitive-center features separately within the depth and RGB modalities.

Let $\boldsymbol{f}_I^\ell(\boldsymbol{x})$ and $\boldsymbol{f}_G^\ell(\boldsymbol{x})$ denote feature vectors bilinearly sampled from the corresponding modality-specific encoder maps. We apply the same sampling at Gaussian centers $\boldsymbol{\mu}_{\ell,i}$ and compute the depth and RGB affinities separately:
\begin{equation}
\begin{aligned}
b_{\ell,i}^{I}(\boldsymbol{x})&=
\exp\!\left[-\frac{\|\boldsymbol{f}_I^\ell(\boldsymbol{x})-
\boldsymbol{f}_I^\ell(\boldsymbol{\mu}_{\ell,i})\|_2^2}
{2(\tau_I^\ell)^2}\right],\\[4pt]
b_{\ell,i}^{G}(\boldsymbol{x})&=
\exp\!\left[-\frac{\|\boldsymbol{f}_G^\ell(\boldsymbol{x})-
\boldsymbol{f}_G^\ell(\boldsymbol{\mu}_{\ell,i})\|_2^2}
{2(\tau_G^\ell)^2}\right].
\end{aligned}
\label{eq:feature_affinities}
\end{equation}
The positive, learnable bandwidths $\tau_I^\ell$ and $\tau_G^\ell$ control sensitivity to feature differences in the two modalities. We combine these affinities with the spatial weight and contribution strength to obtain the aggregation weight:
\begin{equation}
w_{\ell,i}(\boldsymbol{x})=
\alpha_{\ell,i}\,b_{\ell,i}^{\mathrm{geo}}(\boldsymbol{x})
\,b_{\ell,i}^{I}(\boldsymbol{x})\,b_{\ell,i}^{G}(\boldsymbol{x}).
\label{eq:joint_weight}
\end{equation}
The depth and RGB affinities jointly modulate spatial aggregation, reducing the influence of primitives whose features are incompatible with those at the query location.

\textbf{Depth aggregation.} Using SAGR spatial weights and SAGA feature affinities, we reconstruct each level at query pixel centers $\boldsymbol{x}_{\ell,p,q}$ through normalized aggregation:
\begin{equation}
I_\ell(\boldsymbol{x})=\frac{\sum_{i\in\mathcal{N}_\ell(\boldsymbol{x})}w_{\ell,i}(\boldsymbol{x})a_{\ell,i}}{\sum_{i\in\mathcal{N}_\ell(\boldsymbol{x})}w_{\ell,i}(\boldsymbol{x})},
\label{eq:depth_aggregation}
\end{equation}
where $\mathcal{N}_\ell(\boldsymbol{x})$ denotes the local set of Gaussian primitives considered at query location $\boldsymbol{x}$. Then, we combine the rendered maps through Eq.~\eqref{eq:multiscale_reconstruction} to obtain the final depth map.

\subsection{Progressive RGB Guidance Training}
\label{sec:scheduling}

Arbitrary-scale super-resolution methods commonly adopt randomly sampled upsampling factors during training. For GDSR, varying these factors changes the amount of structural detail retained in the depth input, while RGB guidance consistently contains rich appearance details. At larger factors, fine RGB structures may have little corresponding evidence in the depth input, making it difficult to establish cross-modal correspondences early in training. We therefore introduce a progressive RGB guidance training strategy that first emphasizes coarse structures shared by both modalities and gradually incorporates finer RGB details. Specifically, we retain random-scale sampling and apply Gaussian blur to RGB guidance with a given probability, progressively reducing the blur strength during training.

At training step $t$, selected samples use blurred guidance $\boldsymbol G^{(t)}=K_{\sigma(t)}*\boldsymbol G$, where $K_{\sigma(t)}$ is a normalized Gaussian blur kernel; other samples use original RGB guidance. The blur bandwidth follows the schedule in Fig.~\ref{fig:framework}(d):
\begin{equation}
\sigma(t)=\sigma_{\max}\left[\max\!\left(1-\frac{t}{T_{\mathrm b}},0\right)\right]^2,
\label{eq:bandwidth_schedule}
\end{equation}
where $\sigma_{\max}$ is the initial blur bandwidth in RGB pixel units, and $T_{\mathrm b}$ marks the end of blur decay. For each sample, the same RGB guidance is used for Gaussian prediction and the RGB affinity in SAGA. After $T_{\mathrm b}$, training continues with unblurred RGB guidance, also used at inference.

\section{Experiments}
\label{sec:experiments}
\begin{figure*}[!t]
\centering
\captionsetup[subfigure]{font=scriptsize,justification=centering,skip=2pt}
\begin{subfigure}[t]{0.139\textwidth}
\centering
\includegraphics[width=\linewidth]{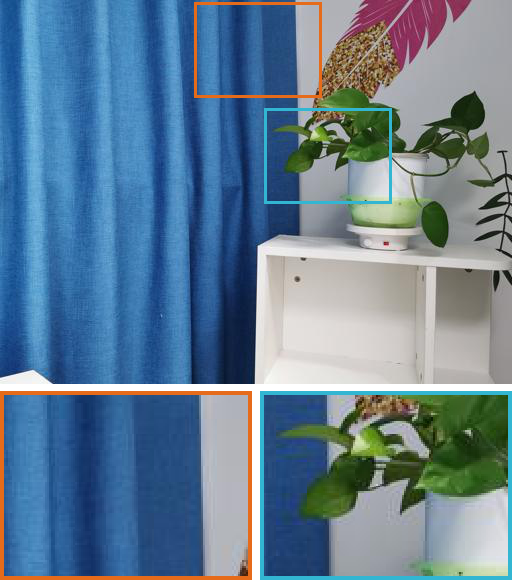}
\caption{Guidance}
\end{subfigure}\hfill
\begin{subfigure}[t]{0.139\textwidth}
\centering
\includegraphics[width=\linewidth]{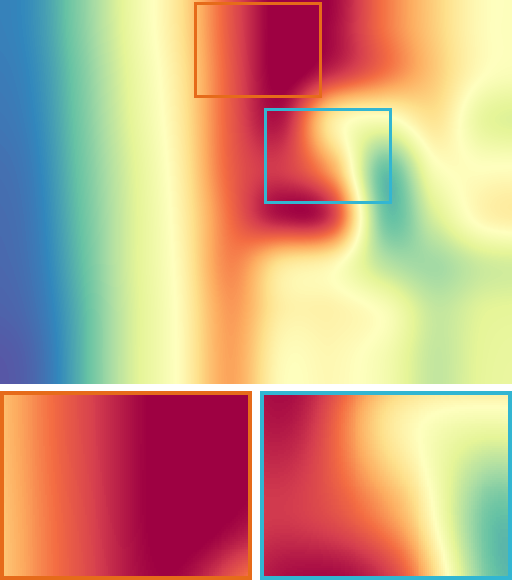}
\caption{Input}
\end{subfigure}\hfill
\begin{subfigure}[t]{0.139\textwidth}
\centering
\includegraphics[width=\linewidth]{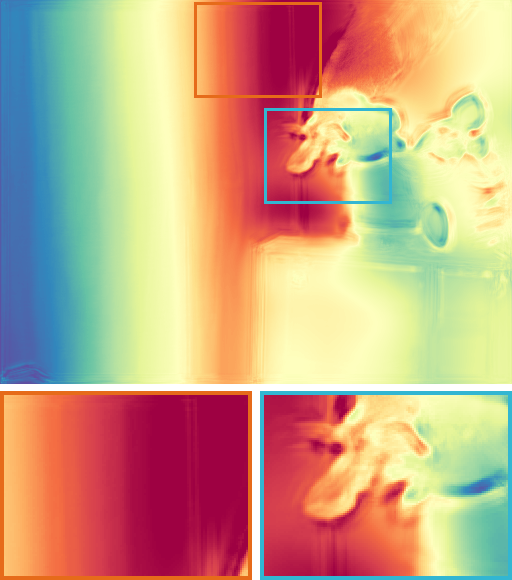}
\caption{DJFR~\cite{li2019djfr}}
\end{subfigure}\hfill
\begin{subfigure}[t]{0.139\textwidth}
\centering
\includegraphics[width=\linewidth]{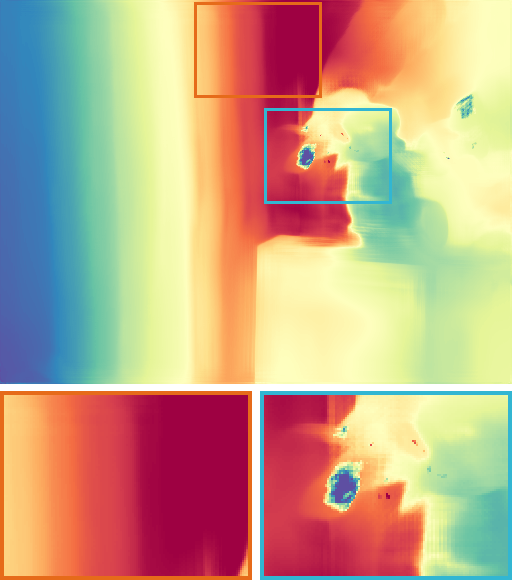}
\caption{DKN~\cite{kim2019dkn}}
\end{subfigure}\hfill
\begin{subfigure}[t]{0.139\textwidth}
\centering
\includegraphics[width=\linewidth]{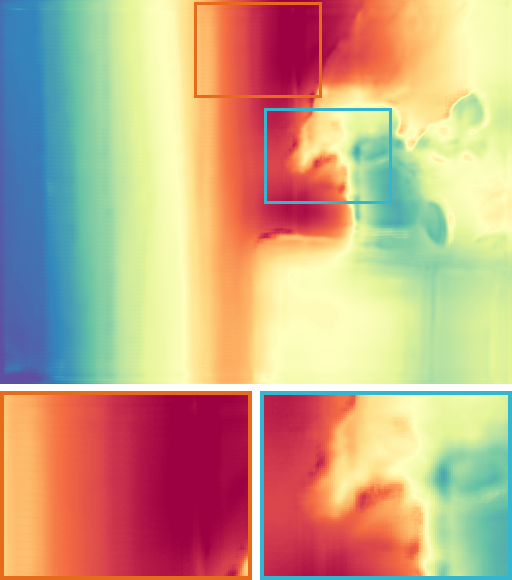}
\caption{DCTNet~\cite{zhao2022dctnet}}
\end{subfigure}\hfill
\begin{subfigure}[t]{0.139\textwidth}
\centering
\includegraphics[width=\linewidth]{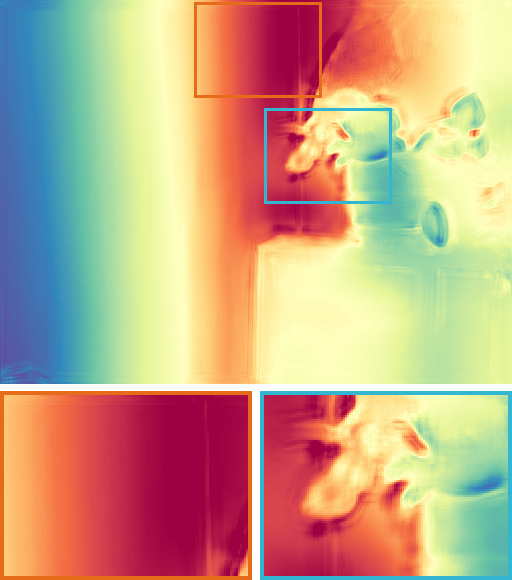}
\caption{SFINet~\cite{zhou2025sfinet}}
\end{subfigure}\hfill
\begin{subfigure}[t]{0.139\textwidth}
\centering
\includegraphics[width=\linewidth]{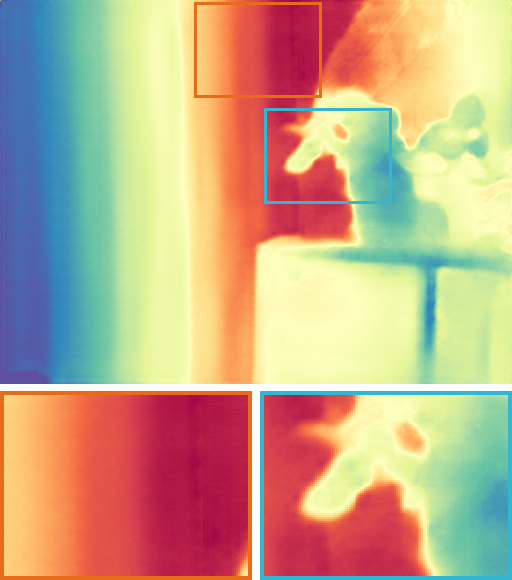}
\caption{GIDF~\cite{zuo2025gidf}}
\end{subfigure}\par\vspace{3pt}
\begin{subfigure}[t]{0.139\textwidth}
\centering
\includegraphics[width=\linewidth]{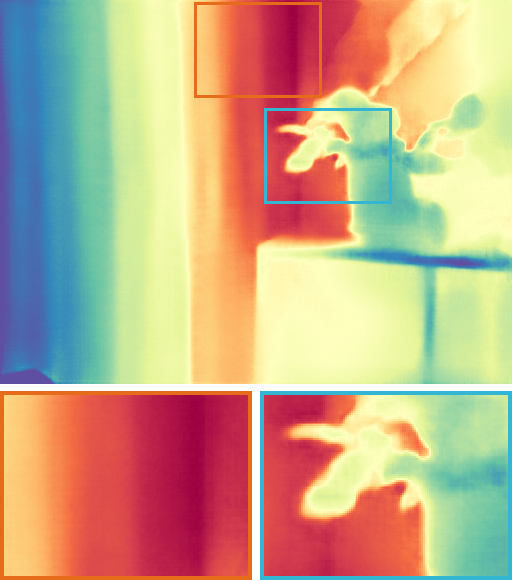}
\caption{GeoDSR~\cite{wang2023geodsr}}
\end{subfigure}\hfill
\begin{subfigure}[t]{0.139\textwidth}
\centering
\includegraphics[width=\linewidth]{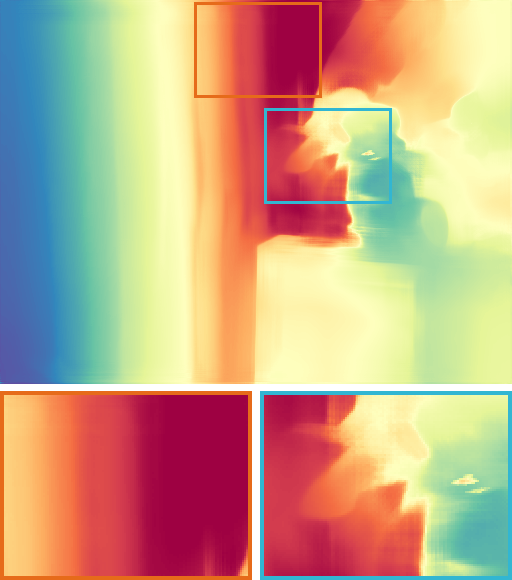}
\caption{SSDNet~\cite{zhao2023ssdnet}}
\end{subfigure}\hfill
\begin{subfigure}[t]{0.139\textwidth}
\centering
\includegraphics[width=\linewidth]{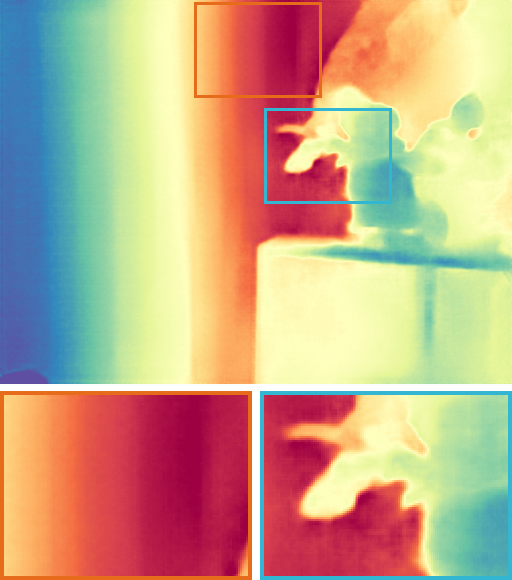}
\caption{DORNet~\cite{wang2025dornet}}
\end{subfigure}\hfill
\begin{subfigure}[t]{0.139\textwidth}
\centering
\includegraphics[width=\linewidth]{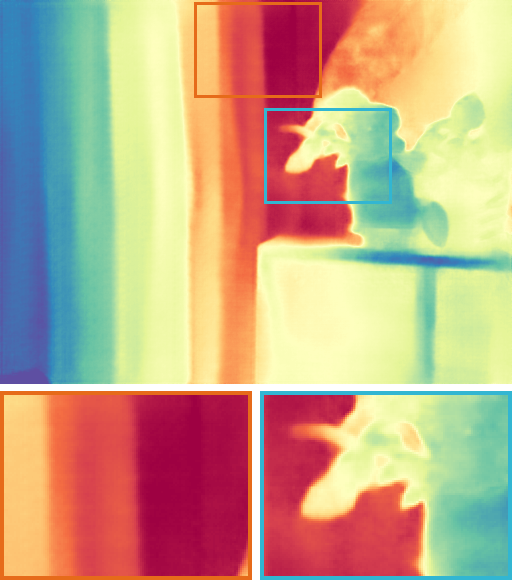}
\caption{DuCos~\cite{yan2025ducos}}
\end{subfigure}\hfill
\begin{subfigure}[t]{0.139\textwidth}
\centering
\includegraphics[width=\linewidth]{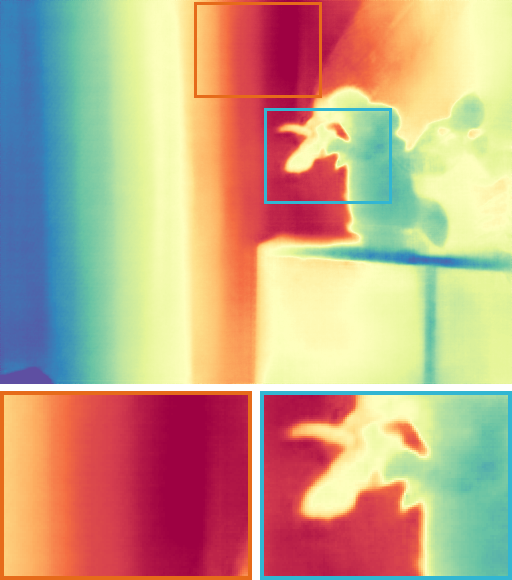}
\caption{LapNet~\cite{zhong2026lapnet}}
\end{subfigure}\hfill
\begin{subfigure}[t]{0.139\textwidth}
\centering
\includegraphics[width=\linewidth]{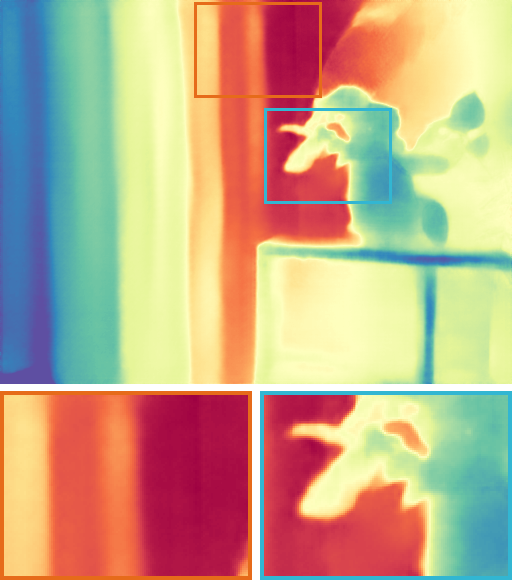}
\caption{\textbf{Ours}}
\end{subfigure}\hfill
\begin{subfigure}[t]{0.139\textwidth}
\centering
\includegraphics[width=\linewidth]{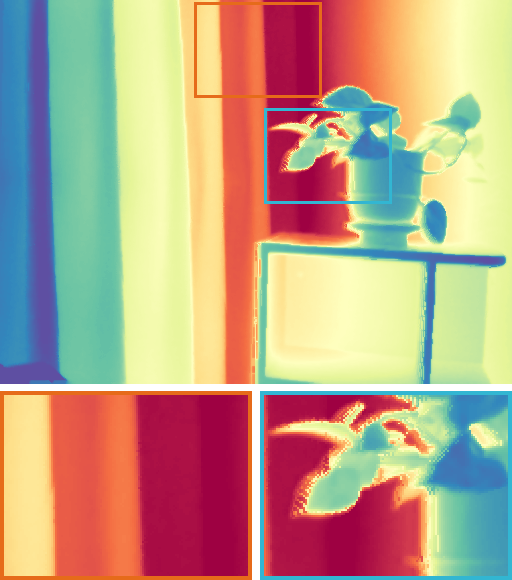}
\caption{GT}
\end{subfigure}
\caption{Visual comparison on RGB-D-D~\cite{he2021rgbdd} at $\times20$. Please enlarge the figure for better visualization.}
\label{fig:supp_rd_x20_062}
\end{figure*}

\begin{table*}[!t]
\centering\footnotesize
\setlength{\tabcolsep}{1.5pt}\renewcommand{\arraystretch}{1.08}
\caption{Quantitative comparison on the RGB-D-D dataset~\cite{he2021rgbdd} for arbitrary-scale RGB-guided depth super-resolution. The best, second-best, and third-best results in each column are highlighted with \colorbox{tablebest}{\strut red}, \colorbox{tablesecond}{\strut orange}, and \colorbox{tablethird}{\strut yellow} backgrounds, respectively.}
\label{tab:rgbdd_metrics}
\begin{tabularx}{\textwidth}{@{}l*{18}{>{\centering\arraybackslash}X}@{}}
\toprule
\multirow{3}{*}{Methods} & \multicolumn{9}{c}{In-distribution} & \multicolumn{9}{c}{Out-of-distribution} \\
\cmidrule(lr){2-10}\cmidrule(lr){11-19}
 & \multicolumn{3}{c}{$\times1$} & \multicolumn{3}{c}{$\times7.5$} & \multicolumn{3}{c}{$\times16$} & \multicolumn{3}{c}{$\times17.5$} & \multicolumn{3}{c}{$\times20$} & \multicolumn{3}{c}{$\times24$} \\
\cmidrule(lr){2-4}\cmidrule(lr){5-7}\cmidrule(lr){8-10}\cmidrule(lr){11-13}\cmidrule(lr){14-16}\cmidrule(lr){17-19}
 & \scriptsize RMSE$\downarrow$ & \scriptsize MAE$\downarrow$ & $\delta_{1.05}\uparrow$ & \scriptsize RMSE$\downarrow$ & \scriptsize MAE$\downarrow$ & $\delta_{1.05}\uparrow$ & \scriptsize RMSE$\downarrow$ & \scriptsize MAE$\downarrow$ & $\delta_{1.05}\uparrow$ & \scriptsize RMSE$\downarrow$ & \scriptsize MAE$\downarrow$ & $\delta_{1.05}\uparrow$ & \scriptsize RMSE$\downarrow$ & \scriptsize MAE$\downarrow$ & $\delta_{1.05}\uparrow$ & \scriptsize RMSE$\downarrow$ & \scriptsize MAE$\downarrow$ & $\delta_{1.05}\uparrow$ \\
\midrule
DJFR~\cite{li2019djfr} & 6.26 & 3.39 & 92.74 & 6.34 & 3.69 & 91.25 & 8.85 & 5.60 & 81.70 & 9.19 & 5.87 & 80.42 & 9.84 & 6.34 & 78.27 & 11.45 & 7.62 & 72.58 \\
SVLRM~\cite{pan2019svlrm} & 5.91 & 3.27 & 93.41 & 6.13 & 3.59 & 91.90 & 8.64 & 5.47 & 82.41 & 8.97 & 5.73 & 81.12 & 9.64 & 6.22 & 78.93 & 11.29 & 7.54 & 72.98 \\
DKN~\cite{kim2019dkn} & 5.56 & 2.96 & 94.77 & 5.66 & 3.23 & 93.50 & 7.72 & 4.80 & 85.91 & 8.06 & 5.07 & 84.53 & 8.78 & 5.59 & 82.02 & 10.60 & 7.03 & 75.13 \\
CUNet~\cite{deng2021cunet} & 5.48 & 2.92 & 94.84 & 5.66 & 3.16 & 93.65 & 7.83 & 4.77 & 85.92 & 8.19 & 5.06 & 84.51 & 8.93 & 5.60 & 81.98 & 10.78 & 7.08 & 74.93 \\
JIIF~\cite{tang2021jiif} & \tabsecond{4.72} & \tabsecond{2.29} & \tabsecond{95.63} & \tabsecond{4.58} & \tabsecond{2.35} & \tabsecond{95.54} & \tabsecond{5.57} & \tabsecond{3.16} & \tabsecond{91.73} & \tabsecond{5.83} & \tabsecond{3.40} & \tabsecond{90.63} & \tabsecond{6.52} & \tabsecond{3.91} & \tabsecond{88.03} & \tabthird{9.00} & \tabthird{5.79} & 79.13 \\
FDSR~\cite{he2021rgbdd} & 5.47 & 2.88 & 94.10 & 5.49 & 2.97 & 93.65 & 7.38 & 4.40 & 86.88 & 7.70 & 4.67 & 85.53 & 8.38 & 5.18 & 83.13 & 10.24 & 6.68 & 75.99 \\
CODON~\cite{yang2022codon} & 5.31 & 2.87 & 95.01 & 5.44 & 3.11 & 93.89 & 7.56 & 4.69 & 86.29 & 7.91 & 4.96 & 84.93 & 8.66 & 5.50 & 82.37 & 10.52 & 6.97 & 75.28 \\
DCTNet~\cite{zhao2022dctnet} & 5.88 & 3.01 & 93.91 & 6.17 & 3.42 & 91.76 & 8.88 & 5.46 & 82.08 & 9.22 & 5.72 & 80.86 & 9.84 & 6.19 & 78.75 & 11.46 & 7.49 & 73.09 \\
MMNet~\cite{yan2022mmnet} & 5.72 & 3.00 & 93.86 & 5.95 & 3.31 & 92.46 & 8.31 & 5.12 & 83.62 & 8.65 & 5.38 & 82.23 & 9.33 & 5.88 & 79.93 & 11.02 & 7.25 & 73.63 \\
SSDNet~\cite{zhao2023ssdnet} & 5.63 & 3.02 & 94.61 & 5.67 & 3.26 & 93.38 & 7.77 & 4.85 & 85.69 & 8.12 & 5.12 & 84.30 & 8.85 & 5.65 & 81.79 & 10.67 & 7.09 & 74.89 \\
GeoDSR~\cite{wang2023geodsr} & 5.01 & 2.45 & 95.35 & 4.87 & 2.54 & 95.01 & 6.00 & 3.47 & 90.70 & 6.31 & 3.73 & 89.52 & 7.08 & 4.29 & 86.70 & 9.68 & 6.25 & 77.67 \\
GIDF~\cite{zuo2025gidf} & 5.12 & 2.50 & 95.07 & 4.96 & 2.55 & 94.92 & 6.02 & 3.44 & 90.90 & 6.29 & 3.67 & 89.86 & 6.95 & 4.17 & 87.45 & 9.02 & 5.81 & \tabthird{79.48} \\
SFINet~\cite{zhou2025sfinet} & 5.71 & 3.01 & 93.87 & 5.94 & 3.31 & 92.47 & 8.33 & 5.13 & 83.56 & 8.67 & 5.40 & 82.17 & 9.34 & 5.89 & 79.88 & 11.02 & 7.25 & 73.61 \\
DORNet~\cite{wang2025dornet} & 5.22 & 2.62 & 94.91 & 5.09 & 2.70 & 94.58 & 6.28 & 3.71 & 89.80 & 6.54 & 3.93 & 88.75 & 7.18 & 4.43 & 86.33 & 9.32 & 6.09 & 78.16 \\
DuCos~\cite{yan2025ducos} & 4.96 & \tabthird{2.44} & 95.35 & 4.79 & \tabthird{2.47} & \tabthird{95.30} & \tabthird{5.76} & \tabthird{3.34} & \tabthird{91.32} & \tabthird{6.03} & \tabthird{3.57} & \tabthird{90.21} & \tabthird{6.68} & \tabthird{4.05} & \tabthird{87.90} & \tabsecond{8.85} & \tabsecond{5.73} & \tabsecond{79.72} \\
LapNet~\cite{zhong2026lapnet} & \tabthird{4.87} & 2.46 & \tabthird{95.44} & \tabthird{4.77} & 2.56 & 95.12 & 5.82 & 3.42 & 91.15 & 6.06 & 3.63 & \tabthird{90.21} & 6.72 & 4.13 & 87.72 & 9.02 & 5.89 & 79.02 \\
\addlinespace[2pt]
\textbf{Ours} & \tabfirst{4.54} & \tabfirst{2.15} & \tabfirst{96.10} & \tabfirst{4.47} & \tabfirst{2.18} & \tabfirst{96.08} & \tabfirst{5.33} & \tabfirst{2.90} & \tabfirst{93.06} & \tabfirst{5.53} & \tabfirst{3.10} & \tabfirst{92.29} & \tabfirst{6.19} & \tabfirst{3.59} & \tabfirst{89.91} & \tabfirst{8.59} & \tabfirst{5.43} & \tabfirst{81.02} \\
\bottomrule
\end{tabularx}
\end{table*}

\subsection{Experimental Settings}
\noindent\textbf{Datasets and protocol.}
We evaluate our method on two widely used real-world datasets: TOFDSR~\cite{yan2025tpvd} and RGB-D-D~\cite{he2021rgbdd}. Both provide paired LR depth maps, GT depth maps, and RGB guidance images. To simulate arbitrary-scale depth super-resolution under real-world conditions, we further downsample the captured LR depth maps, which contain real sensor degradations, using bicubic interpolation. A downsampling factor of 1 retains the original sensor depth, while larger factors further reduce its spatial resolution. We use the officially released training and test splits.

\begin{figure*}[!t]
    \centering
    \captionsetup[subfigure]{font=scriptsize,labelformat=empty,justification=centering,skip=2pt}
    \begin{subfigure}[t]{0.190\textwidth}
        \vspace{0pt}\centering
        \includegraphics[width=\linewidth]{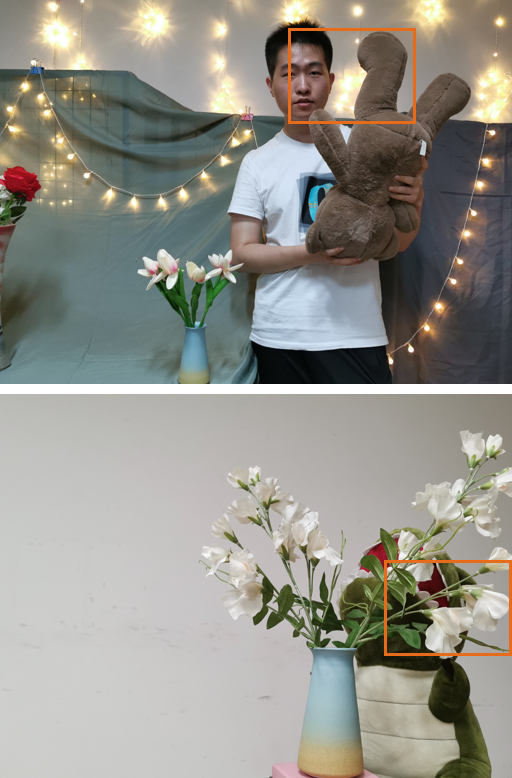}
        \caption{RGB}
    \end{subfigure}\hspace{0.00145\textwidth}\begin{subfigure}[t]{0.095\textwidth}
        \vspace{0pt}\centering
        \includegraphics[width=\linewidth]{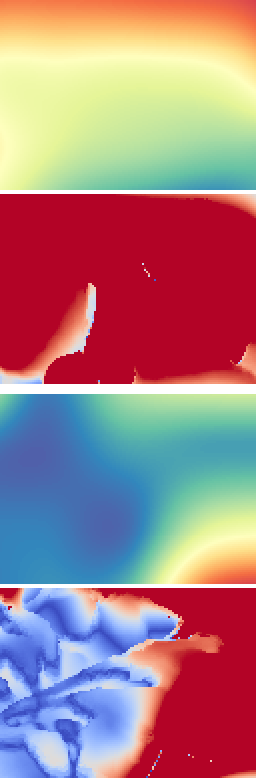}
        \caption{Input}
    \end{subfigure}\hspace{0.00145\textwidth}\begin{subfigure}[t]{0.095\textwidth}
        \vspace{0pt}\centering
        \includegraphics[width=\linewidth]{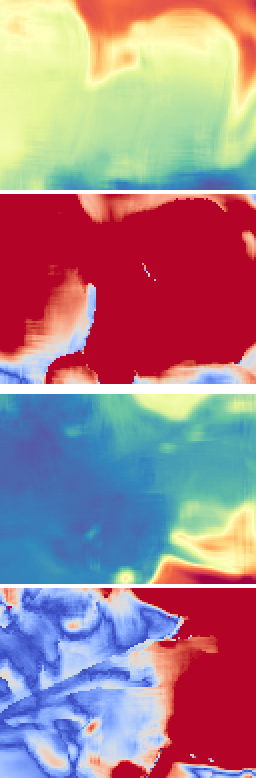}
        \caption{DCTNet~\cite{zhao2022dctnet}}
    \end{subfigure}\hspace{0.00145\textwidth}\begin{subfigure}[t]{0.095\textwidth}
        \vspace{0pt}\centering
        \includegraphics[width=\linewidth]{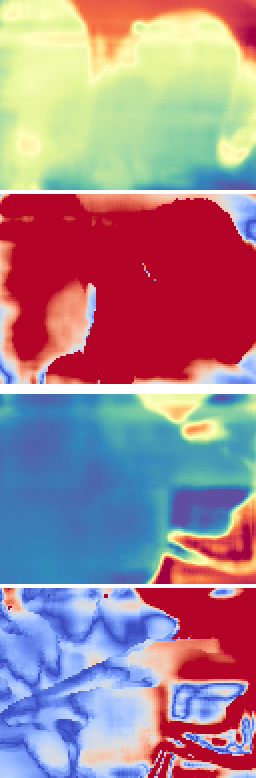}
        \caption{SSDNet~\cite{zhao2023ssdnet}}
    \end{subfigure}\hspace{0.00145\textwidth}\begin{subfigure}[t]{0.095\textwidth}
        \vspace{0pt}\centering
        \includegraphics[width=\linewidth]{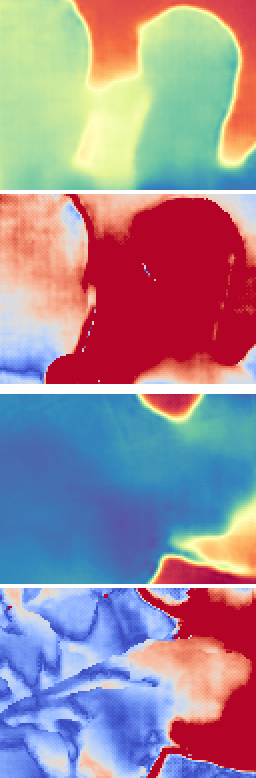}
        \caption{DORNet~\cite{wang2025dornet}}
    \end{subfigure}\hspace{0.00145\textwidth}\begin{subfigure}[t]{0.095\textwidth}
        \vspace{0pt}\centering
        \includegraphics[width=\linewidth]{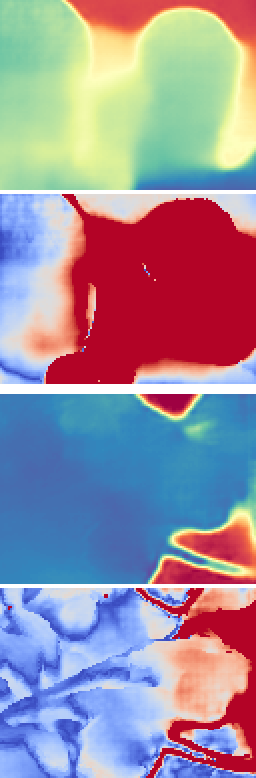}
        \caption{DuCos~\cite{yan2025ducos}}
    \end{subfigure}\hspace{0.00145\textwidth}\begin{subfigure}[t]{0.095\textwidth}
        \vspace{0pt}\centering
        \includegraphics[width=\linewidth]{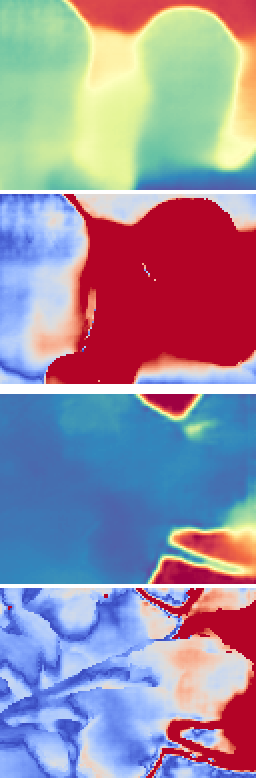}
        \caption{LapNet~\cite{zhong2026lapnet}}
    \end{subfigure}\hspace{0.00145\textwidth}\begin{subfigure}[t]{0.095\textwidth}
        \vspace{0pt}\centering
        \includegraphics[width=\linewidth]{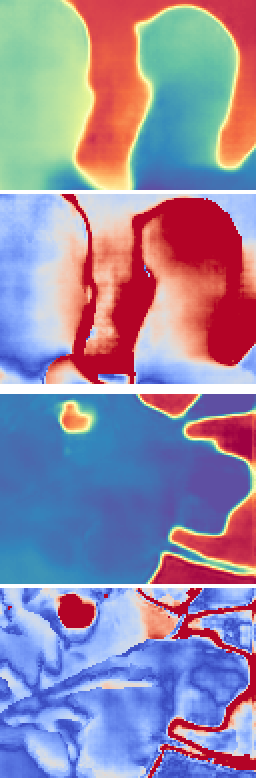}
        \caption{\textbf{Ours}}
    \end{subfigure}\hspace{0.00145\textwidth}\begin{subfigure}[t]{0.095\textwidth}
        \vspace{0pt}\centering
        \includegraphics[width=\linewidth]{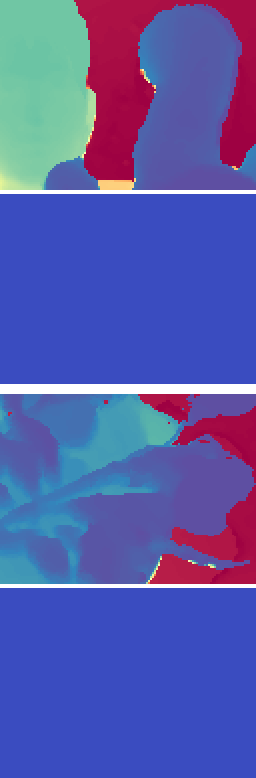}
        \caption{GT}
    \end{subfigure}\hspace{0.00145\textwidth}\begin{minipage}[t]{0.03673828125\textwidth}
        \vspace{0pt}\includegraphics[width=\linewidth]{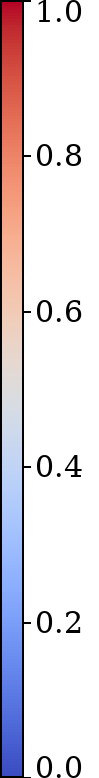}
    \end{minipage}
    \caption{Visual comparison of reconstructed depth maps and error maps on TOFDSR~\cite{yan2025tpvd} at $\times16$. Please zoom in for a better view.}
    \label{fig:qualitative_tofdsr}
\end{figure*}

\begin{figure*}[t]
\centering
\captionsetup[subfigure]{font=footnotesize,labelfont=normalfont,justification=centering,skip=3pt}
\begin{subfigure}[b]{0.242\textwidth}
\centering\includegraphics[width=\linewidth]{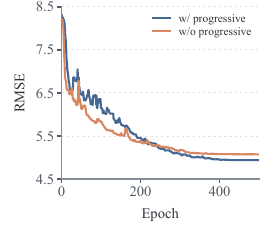}
\caption{Progressive RGB guidance}\label{fig:progressive_rgb_guidance}
\end{subfigure}\hfill
\begin{subfigure}[b]{0.242\textwidth}
\centering\includegraphics[width=\linewidth]{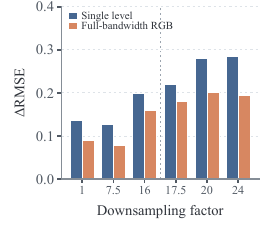}
\caption{Hierarchy and training}\label{fig:component_hierarchy}
\end{subfigure}\hfill
\begin{subfigure}[b]{0.242\textwidth}
\centering\includegraphics[width=\linewidth]{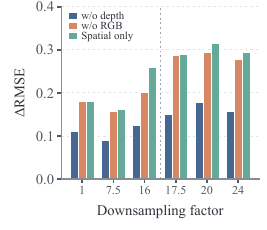}
\caption{Cross-modal affinities}\label{fig:component_affinity}
\end{subfigure}\hfill
\begin{subfigure}[b]{0.242\textwidth}
\centering\includegraphics[width=\linewidth]{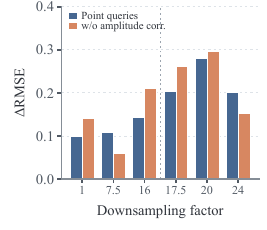}
\caption{Pixel-footprint rendering}\label{fig:component_footprint}
\end{subfigure}\hfill
\vspace{-0.1in}
\caption{Training and component analysis on RGB-D-D. (a) Validation RMSE at $\times16$ with and without progressive RGB guidance training. (b) Gaussian hierarchy and RGB bandwidth scheduling. (c) Depth and RGB affinities in SAGA. (d) Pixel footprint and amplitude correction in SAGR. Panels (b--d) show RMSE increases relative to the full model across downsampling factors.}\label{fig:rgbdd_ablation_preview}
\vspace{-0.1in}
\end{figure*}

\noindent\textbf{Implementation details.}
Our framework is implemented in PyTorch. We train one model per dataset on NVIDIA GeForce RTX 5090 GPUs for 500 epochs using Adam, an $L_1$ reconstruction loss, and a batch size of 16. During training, we uniformly sample an additional downsampling factor from $[1,16]$ for each sample. Evaluation covers $\times1$, $\times7.5$, and $\times16$ within the training range, and $\times17.5$, $\times20$, and $\times24$ beyond it. The learning rate follows a cosine annealing schedule from 1e-4 to 1e-6. Following DuCos~\cite{yan2025ducos}, we report root mean squared error (RMSE), mean absolute error (MAE), and depth threshold accuracy $\delta_{1.05}$. Lower RMSE and MAE and higher $\delta_{1.05}$ indicate better performance. Further implementation details and metric definitions are provided in the \textbf{Supplementary Material}.

\subsection{Comparison with Existing Methods}
\noindent\textbf{Compared methods.} We compare with sixteen learning-based GDSR methods. Among them, GeoDSR~\cite{wang2023geodsr}, JIIF~\cite{tang2021jiif}, and GIDF~\cite{zuo2025gidf} employ implicit representations to support arbitrary-scale reconstruction. For a fair comparison, we retrain all competing methods using their official implementations under a unified protocol. All methods share the same dataset splits, LR input preparation, and evaluation metrics. During training, additional downsampling factors are sampled uniformly from $[1,16]$. Each method uses one model per dataset across all test factors.

\noindent\textbf{Quantitative comparison.}
Tables~\ref{tab:tof_metrics} and~\ref{tab:rgbdd_metrics} present quantitative results on TOFDSR and RGB-D-D, respectively. Our method achieves the best performance across all three metrics at every evaluated factor on both datasets. This advantage extends to large downsampling factors beyond the training range, where sparse depth observations make reconstruction more challenging. These results demonstrate effective reconstruction across varying input resolutions and generalization to unseen downsampling factors.

\noindent\textbf{Computational efficiency.}
We further compare reconstruction accuracy and model complexity, as shown in Fig.~\ref{fig:efficiency}. Our method achieves a favorable trade-off between accuracy and model size, delivering more accurate depth reconstruction with fewer parameters than JIIF and DuCos.

\noindent\textbf{Qualitative comparison.}
Figures~\ref{fig:supp_rd_x20_062} and~\ref{fig:qualitative_tofdsr} show reconstructed depth maps on RGB-D-D at $\times20$ and TOFDSR at $\times16$, respectively. On RGB-D-D, the enlarged curtain and plant regions illustrate differences in boundary localization and the recovery of thin structures. On TOFDSR, the boundaries around the head, toy, and flowers reveal the effects of sparse depth sampling, while the corresponding error maps expose deviations that are less apparent in the depth colors. Our results preserve object contours and reduce errors in the highlighted boundary regions.

\begin{figure*}[!t]
    \centering
    \captionsetup[subfigure]{font=footnotesize,labelfont=normalfont,justification=centering,skip=3pt}
    \begin{subfigure}[t]{0.30\textwidth}
        \vspace{0pt}
        \centering
        \includegraphics[width=\linewidth]{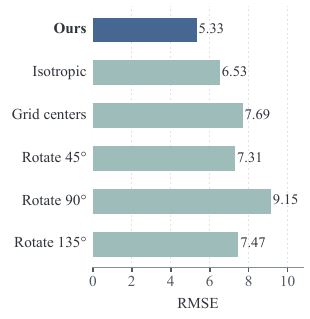}
        \caption{Overall RMSE at $\times16$}
        \label{fig:geometry_diagnostic}
    \end{subfigure}\hspace{0.008\textwidth}\begin{subfigure}[t]{0.688\textwidth}
        \vspace{0pt}
        \centering
        \includegraphics[width=\linewidth]{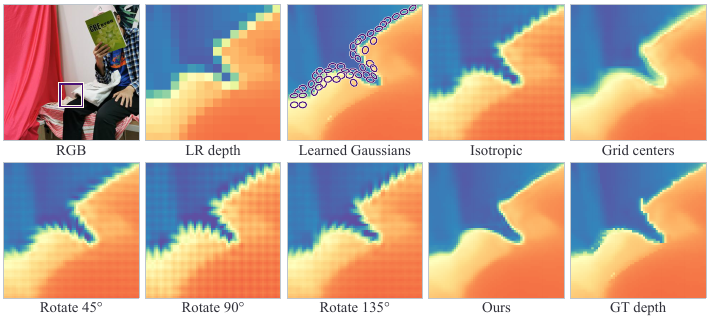}
        \caption{Visual comparison under Gaussian parameter variations}
        \label{fig:gaussian_analysis2}
    \end{subfigure}
    \vspace{-0.1in}
    \caption{Effect of learned Gaussian geometry on depth reconstruction on RGB-D-D at $\times16$. (a) RMSE after perturbing Gaussian shapes, centers, and orientations. (b) Learned Gaussian distributions and local reconstructions under the corresponding perturbations.}
    \label{fig:gaussian_guidance_analysis}
    \vspace{-0.1in}
\end{figure*}

\subsection{Analysis of Model Components}
We conduct ablation studies on RGB-D-D to evaluate progressive RGB guidance, the hierarchical Gaussian representation, cross-modal affinities in SAGA, and pixel-footprint rendering in SAGR. Component variants are compared with the full model across downsampling factors within and beyond the training range, using RMSE as the evaluation metric. We also assess learned Gaussian geometry by perturbing kernel shapes, centers, and orientations at inference.

\noindent\textbf{Progressive RGB guidance.}
We compare progressive RGB guidance with full-bandwidth guidance throughout training. As shown in Fig.~\ref{fig:progressive_rgb_guidance}, the progressive variant initially exhibits slightly higher validation RMSE because blurred guidance provides fewer fine structural cues. However, it eventually outperforms the full-bandwidth baseline as training progresses. This suggests that suppressing fine RGB textures early in training helps the model first learn coarse structures shared by RGB and depth, with less interference from appearance variations that do not correspond to depth changes. As the blur decreases, this structural foundation supports more effective use of finer RGB details, leading to lower final reconstruction error.

\noindent\textbf{Hierarchical Gaussian representation.}
We replace the hierarchical representation with a single Gaussian level to assess the role of hierarchical reconstruction. Figure~\ref{fig:component_hierarchy} shows that this replacement increases RMSE at every evaluated factor, with larger increases beyond the training range. This supports combining Gaussian fields across levels for depth reconstruction at varying input resolutions.

\noindent\textbf{Cross-modal affinities.}
We evaluate three SAGA variants: removing depth affinity, removing RGB affinity, and using spatial weights alone. Both modalities remain available for Gaussian prediction. As shown in Fig.~\ref{fig:component_affinity}, removing either affinity increases RMSE, while spatial weighting alone produces the largest increase at every factor. Joint depth and RGB affinities thus improve reconstruction over aggregation based solely on local spatial proximity.

\noindent\textbf{Pixel-footprint rendering.}
We evaluate SAGR using point queries without footprint convolution and footprint-aware kernels without amplitude correction. The latter retains covariance broadening but removes the integral-preserving amplitude factor. Both variants increase RMSE across all evaluated factors in Fig.~\ref{fig:component_footprint}. These results support accounting for the output pixel's spatial coverage and preserving the kernel integral, with both components contributing to reconstruction within and beyond the training range.

\noindent\textbf{Gaussian parameter sensitivity.}
\label{sec:learned_gaussian_analysis}
We assess the role of learned Gaussian geometry by replacing anisotropic kernels with isotropic ones, resetting centers to grid locations, and rotating kernel orientations at inference. As shown in Fig.~\ref{fig:geometry_diagnostic}, all tested perturbations increase RMSE on RGB-D-D at $\times16$. Isotropic kernels and grid-reset centers raise RMSE from 5.33 to 6.53 and 7.69, respectively, while a $90^\circ$ rotation yields the largest reconstruction error of 9.15.

Figure~\ref{fig:gaussian_analysis2} shows the corresponding local reconstructions at $\times16$. The learned Gaussians follow the depth contour with varying shapes and orientations. Isotropic kernels introduce irregular boundaries and surface ripples, while resetting centers removes fine contour details. Kernel rotations produce jagged edges and surface artifacts. The learned geometry preserves a clearer boundary and smoother surface interiors. Together, these observations suggest that adaptive Gaussian support follows local depth structure, promoting aggregation within surfaces while preserving depth discontinuities along object boundaries.

\section{Conclusion}
In this paper, we propose a cross-modal Gaussian rendering framework for arbitrary-scale guided depth super-resolution. The framework learns hierarchical Gaussian representations that support continuous depth reconstruction. Structure-aware aggregation combines depth and RGB feature affinities to preserve depth boundaries, while scale-adaptive rendering accounts for the target pixel footprint. A progressive RGB guidance strategy further improves reconstruction by gradually introducing finer RGB details during training. Experiments on two real-world benchmarks demonstrate consistent improvements over competing methods at scales within and beyond the training range, while maintaining a favorable balance between reconstruction accuracy and computational efficiency.

\section*{LLM Usage Statement}
During the preparation of this manuscript, DeepSeek was utilized to help translate the text into English, and refine the academic language. The model was not used for model development, data synthesis, or image manipulation. All experimental analyses and conclusions were verified by the authors, who take full responsibility for the final content.

\FloatBarrier
\end{document}